\documentclass{dynafront2026} 

\usepackage{caption}
\usepackage{placeins}

\usepackage[dvipsnames]{xcolor}
\definecolor{aaroncolor}{RGB}{101,67,30}
\definecolor{sylvaincolor}{RGB}{0,0,139}

\title[Controllable Diverse Sampling through Pullback Directions]
      {Controllable Diverse Sampling through Pullback Directions}

\optauthor{%
    \begin{tabular}{@{}c@{}}
        \begin{tabular*}{\textwidth}{@{\extracolsep{\fill}}cccc@{}}
            \Name{Sylvain {Dehayem Kenfouo}}%
                {\normalfont
                    \begin{NoHyper}\thanks{These authors contributed equally.}\end{NoHyper}%
                    \textsuperscript{1}}
            &
            \Name{Aaron Zweig}%
            \textsuperscript{\normalfont *2,3}
            &
            \Name{Mingxuan Zhang}%
            \textsuperscript{\normalfont 3,4}
            &
            \Name{Elham Azizi}%
            \textsuperscript{\normalfont 3,4,5,6,7}
        \end{tabular*}\\[1em]
        {\normalfont\normalsize
            \begin{tabular}{@{}c@{}}
                \textsuperscript{1}\'Ecole Polytechnique, Institut Polytechnique de Paris\\
                \textsuperscript{2}New York Genome Center\\
                \textsuperscript{3}Irving Institute for Cancer Dynamics, Columbia University\\
                \textsuperscript{4}Department of Systems Biology, Columbia University\\
                \textsuperscript{5}Department of Computer Science, Columbia University\\
                \textsuperscript{6}Department of Biomedical Engineering, Columbia University\\
                \textsuperscript{7}Data Science Institute, Columbia University
            \end{tabular}}
    \end{tabular}%
}

\makeatletter
\renewcommand{\@titlefoot}{}
\makeatother

\begin{document}

\maketitle

\begin{abstract}%
% Specify abstract that summarizes the content of the paper.

Conditional diffusion models can generate high-quality images that align closely with a given prompt. However, samples generated from different random seeds may remain visually similar, limiting creative exploration. Existing approaches can increase variation, but often require additional optimization or sampling overhead, while stronger perturbations may introduce distortions that degrade image quality or prompt alignment.  We propose a training-free method that leverages the local pullback geometry induced by the mapping from the conditioning-embedding space to the image space to identify directions of maximal output sensitivity. By steering different sampling trajectories along these directions, we encourage distinct generations while maintaining consistency with the conditioning signal. Experiments on MS-COCO and BrushNet show the efficacy of the method in improving diversity on prompt-conditioned image generation and inpainting.

\end{abstract}

%\begin{keywords}%
%  List of keywords%
%\end{keywords}

% \section{Main}

% % \begin{itemize}
% %     \item Introduce the diffusion model (One approach) and show the relation with the image space
% %     \item talk about conditional generation, and Diversity generation, 
% %     \item Introduce the classifier Guidance as key guidance of the model generation
% %     \item Introduce the pullback 
% %     \item Talk about the method
% %     \item Talk about inpainting 
% %     \item Present the experiment
% %     \item 
% % \end{itemize}
% Main text limited to 9 pages, excluding references and appendices.
\section{Introduction}

Recent advances in diffusion models \citep{ho2020denoising,song2019generative,song2020denoising,song2020score,rombach2022high}, have
enabled text-to-image systems to generate visually compelling images that
closely follow natural-language prompts. However, image quality and conditional alignment do not fully characterize a generative model. For a fixed condition, the model should also represent diverse valid outputs. 

In practice, trajectories may concentrate
around the same dominant modes of the conditional distribution, resulting in
semantically redundant outputs. This limited coverage increases the cost of
obtaining a diverse set of samples.  Inverse problems like image inpainting\citep{ju2024brushnet} further constrain diversity, as only a small masked region is generated.

Many methods consider postprocessing steps that require many more gradient calculations, typically fine-tuning either the score model or the initial noise distribution to optimize a measure of diversity.  Alternatively, inference-time methods based on sequential Monte Carlo can introduce diversity by drawing many samples and iteratively resampling to enforce a final distribution with high diversity.  Both these methods suffer from substantially increased time complexity, either during fine-tuning or during inference.  In this work, we therefore seek to answer:

\begin{center}
    \textit{How can a pretrained diffusion model generate high-quality, diverse samples without fine tuning or sampling prohibitively many particles?}
\end{center}

We propose an efficient guidance method to encourage diverse samples.  Namely, for a given prompt and possible inpainting mask, we require only a constant number of Jacobian evaluations (as few as one) from which we can generate an arbitrary number of diverse samples that are conditioned by the text and/or mask.  We demonstrate improved performance on diversity metrics over comparable baselines operating in the regime of no fine-tuning and cheap inference.

\section{Preliminaries}

\paragraph{Diffusion Models.}
\label{sec:ddpm}

We consider Denoising Diffusion Probabilistic Models (DDPM), which define a forward process that gradually corrupts samples from the data distribution with Gaussian noise \citep{ho2020denoising}. Given a clean image $x_0 \sim p_{\mathrm{data}}$, the forward Markov chain is defined as
\begin{equation}
    q(x_{1:T}\mid x_0)
    =
    \prod_{t=1}^{T} q(x_t\mid x_{t-1}),
    \qquad
    q(x_t\mid x_{t-1})
    =
    \mathcal{N}
    \left(
        x_t;
        \sqrt{\alpha_t}\,x_{t-1},
        (1-\alpha_t)I
    \right),
\end{equation}
where $\alpha_t=1-\beta_t$ and $\{\beta_t\}_{t=1}^{T}$ denotes a predefined noise schedule.  We let $\bar{\alpha}_t=\prod_{s=1}^{t}\alpha_s$.
% Letting
% $\bar{\alpha}_t=\prod_{s=1}^{t}\alpha_s$, the marginal distribution at an arbitrary timestep admits the closed form
% \begin{equation}
%     q(x_t\mid x_0)
%     =
%     \mathcal{N}
%     \left(
%         x_t;
%         \sqrt{\bar{\alpha}_t}\,x_0,
%         (1-\bar{\alpha}_t)I
%     \right),
% \end{equation}.
For a sufficiently long diffusion process, $x_T$ approaches an isotropic Gaussian distribution. Given a network $\epsilon_\theta$ trained to predict the noise used to generate $x_t$ conditioned on the condition $c$, the reverse process is modeled by
\begin{equation}
    p_\theta(x_{t-1}\mid x_t,c)
    =
    \mathcal{N}
    \left(
        x_{t-1};
        \frac{1}{\sqrt{\alpha_t}}
        \left(
            x_t
            -
            \frac{\beta_t}{\sqrt{1-\bar{\alpha}_t}}
            \epsilon_\theta(x_t,t,c)
        \right),
        \sigma_t^2 I
    \right).
\end{equation}

% Under the Gaussian
% forward process, the noise predictor is related to the conditional score
% through
% \begin{equation}
%     s_\theta(x_t,t,c)
%     \approx
%     \nabla_{x_t}\log p_t(x_t\mid c)
%     =
%     -\frac{\epsilon_\theta(x_t,t,c)}
%     {\sqrt{1-\bar{\alpha}_t}}.
%     \label{eq:score_noise_relation}
% \end{equation}

The predicted noise can then be used to estimate the original clean image
at any timestep. Using Tweedie's formula, the denoised estimate is
\begin{equation}
    \widehat{x}_{0\mid t}(x_t,t,c)
    := \mathbb{E}[x_0\mid x_t,c] =
    \frac{1}{\sqrt{\bar{\alpha}_t}}
    \left(
        x_t
        -
        \sqrt{1-\bar{\alpha}_t}\,
        \epsilon_\theta(x_t,t,c)
    \right).
    \label{eq:tweedie_estimate}
\end{equation}
% or, equivalently,
% \begin{equation}
%     \widehat{x}_{0\mid t}(x_t,t,c)
%     =
%     \frac{1}{\sqrt{\bar{\alpha}_t}}
%     \left(
%         x_t
%         +
%         (1-\bar{\alpha}_t)s_\theta(x_t,t,c)
%     \right).
% \end{equation}
It therefore defines a differentiable
mapping from the current noisy state to an estimate of the final clean
image, which will later be used to construct the pullback geometry.

% \subsection{Classifier-Free Guidance and Image Inpainting}
% \label{sec:guidance_inpainting}

% \paragraph{Classifier-free guidance.}
% Conditional diffusion models generate samples according to side information
% $c$, such as a text embedding. Classifier-free guidance (CFG) strengthens
% the influence of this condition during sampling by combining conditional and
% unconditional noise predictions \citep{ho2022classifier}:
% \begin{equation}
%     \epsilon_{\mathrm{cfg}}(x_t,t,c)
%     =
%     \epsilon_\theta(x_t,t,\varnothing)
%     +
%     w
%     \left(
%         \epsilon_\theta(x_t,t,c)
%         -
%         \epsilon_\theta(x_t,t,\varnothing)
%     \right),
%     \label{eq:cfg}
% \end{equation}
% where $\varnothing$ denotes the null condition and $w\geq 1$ is the
% guidance scale. The difference
% \begin{equation}
%     \epsilon_\theta(x_t,t,c)
%     -
%     \epsilon_\theta(x_t,t,\varnothing)
% \end{equation}
% captures the direction induced by the condition, while $w$ controls its
% strength. Increasing the guidance scale generally improves alignment with
% $c$, but can reduce the coverage of the conditional distribution.

\paragraph{Inpainting as an inverse problem.}
In image restoration problems, only a partial or corrupted measurement
$y$ of image $x_0$ is available.  For noiseless image inpainting, let
$M\in\{0,1\}^d$ denote the mask of the observed region, then $y=M\odot x_0$.  Text guidance controls the semantic content through $c$, whereas the
mask constrains the generated sample to agree
with the visible region $y$.

% The observation process can be
% written
% \begin{equation}
%     y
%     =
%     \mathcal{A}(x_0)+n,
%     \label{eq:inverse_problem}
% \end{equation}
% where $\mathcal{A}:\mathbb{R}^d\rightarrow\mathbb{R}^m$ is a possibly
% non-invertible measurement operator and $n$ denotes measurement noise \citep{chung2022diffusion}.
% When $\mathcal{A}$ is many-to-one, recovering $x_0$ from $y$ defines an
% ill-posed inverse problem. 
% In a Bayesian formulation, the diffusion model
% provides a prior over images, and reconstruction amounts to sampling from
% the posterior distribution
% \begin{equation}
%     p(x_0\mid y,c)
%     \propto
%     p(y\mid x_0,c)\,p(x_0\mid c).
% \end{equation}

\section{Conditional Pullback Metric}
In this section, we introduce a differential geometric framework for
identifying directions in the conditioning space of
diffusion models.  In Euclidean spaces with a smooth map $f: \mathbb{R}^n \rightarrow \mathbb{R}^m$, the pullback metric at $x \in \mathbb{R}^n$ evaluated at $v \in \mathbb{R}^n$ is defined as $\|J_{f(x)} v\|$.  We therefore seek directions in the input space that induce proportionally larger shifts in the output space via the top right singular vectors of $J_{f(x)}$.  We consider the Jacobian of the noise predictor with respect to condition:

\begin{equation}
J_c(x_t, t, c) = \frac{\partial\epsilon_\theta(x_t,t,c)}{\partial c},
\end{equation}

% Therefore, the squared Euclidean distance induced in the response space is
% \begin{equation}
% \boxed{
% \|v\|_{\mathrm{pb},C}^{2}
% \triangleq
% \langle u,u\rangle_{\mathcal Y}
% =
% v^\top J_C^\top J_Cv,
% \qquad
% u=J_Cv
% }
% \label{eq:conditional_pullback_norm}
% \end{equation}
% we can compute the pullback metric direction
% \paragraph{Image inpainting}
In the case of image inpainting, we can chain the linear masking process with Tweedie's formula to define a Jacobian from the context to the masked portion of the image.  Let $M_{edit}=1-M$ denote the bit-flipped mask of $M$, then we can define the linear operator $W_M$ acting on image space by 
$$
W_M v
=
M_{edit} \odot v,
\label{eq:latent_mask_operator}
$$
The differential of the
masked clean estimate is

\begin{equation}\label{eq:masked_clean_jacobian}
\frac{
\partial
\left(
W_M\widehat{x}_{0\mid t}(x_t, t, c)
\right)
}{
\partial c
}
=
-
\sqrt{
\frac{1-\bar{\alpha}_t}{\bar{\alpha}_t}
}
W_MJ_c(x_t, t, c).
\end{equation}
% Then, the regional pullback norm is

% \begin{equation}
% \boxed{
% \|v\|_{\mathrm{pb},C,M}^{2}
% \triangleq
% \langle u_M,u_M\rangle_{\mathcal Y}
% =
% v^\top
% J_C^\top W_M^\top W_MJ_C
% v.
% }
% \label{eq:regional_pullback_norm}
% \end{equation}

% The factor
% $(1-\bar{\alpha}_t)/\bar{\alpha}_t$ is and independent of $v$ and doesn't change the eigenvectors. We can
% consequently compute the regional pullback metric directly from the predicted
% noise response:

% \begin{equation}
% \boxed{
% G_{C,M}
% =
% J_C^\top W_M^\top W_MJ_C.
% }
% \label{eq:regional_pullback_metric}
% \end{equation}

Using the conditional pullback metric, we can infer a subspace that locally produces large variability in the latent space.  Explicitly, we consider span of the top $K$ right singular vectors of Jacobian $ \widetilde J_c=W_MJ_c$.  Since explicitly constructing the condition Jacobian is computationally
intractable, we use a sketching method to obtain these directions using only Jacobian-vector products
dominant singular directions~\cite{halko2011finding}, for more details see Section~\ref{sec:approximate-svd}.

\section{Pullback-informed prompt perturbation}

% A core challenge in text-to-image diffusion models is that independently sampled initial noises $z_T$ may still produce generations that converge toward semantically similar modes, despite the stochasticity of the sampling process. We act on the reverse trajectory by introducing small perturbations during denoising. 

% \citep{park2023understanding} introduces pullback metrics to characterize the local geometry of diffusion latent spaces by relating them to a representation space with a known metric. This construction is particularly relevant in our setting because it provides a principled way to transfer geometric information between two spaces. 

% Experiments on figure \ref{fig:token_cross_attention} and \citep{gal2022image}, demonstrates the strong influence of prompt embeddings on the generated image. We consider the mapping induced by the conditional diffusion model from the condition space $\mathcal C$ to the image space $\mathcal Y$. Since text-to-image diffusion models aim to sample from $p(y|c)$ \citep{ho2022classifier}, modifying the condition associated with a fixed noisy trajectory can lead to substantially different samples.

% At each timestep $t$, and for a fixed noisy latent $z_t$, the conditional
% denoising network defines the map
% \[
% f_t:\mathcal{C}\rightarrow\mathcal{Z},
% \qquad
% C\mapsto\epsilon_\theta(z_t,t,C),
% \]
% from the prompt-embedding space $\mathcal{C}$ to the predicted-noise space
% $\mathcal{Z}$. 

With the top singular vectors, we can identify directions in the prompt embedding space along which small perturbations induce large changes in the denoising prediction and, consequently, in the predicted clean latent.  We use these directions to diversify the reverse trajectories while preserving the semantic content of the original condition.

\paragraph{Initialization and trajectory diversification.}

We first sample a batch of $B$ independent initial latent variables, $\mathcal Z_T=\{z_T^{(i)}\}_{i=1}^{B}$ with $z_T^{(i)}\sim\mathcal N(0,I),$
and use the same latent batch for all methods considered in our experiments.  Given the base prompt $c_0$, a pullback basis is then estimated from an anchor trajectory at the reference
timestep $t_b$. The dominant eigenvectors of the condition-space metric
$G_{t_b}=J_{c_0,t_b}^{\top}J_{c_0,t_b}$ are partitioned into mutually disjoint
subspaces, one for each particle. Each particle is assigned a persistent direction obtained by projecting a fixed Gaussian field onto its corresponding subspace. The perturbations are mutually orthogonal and encourage different local variations of the conditional model. 
% The directions are rescaled according to the norm of the clean
% prompt embedding and are applied only to the real prompt tokens.
For particle $i$, the perturbed condition is therefore written as
$$
c_i(t)=c_0+\rho_i^\star\gamma_t\Delta c_i,
$$
where $\Delta c_i$ denotes the prompt-relative pullback direction
defined in Equation \ref{eq:delta_ci}, $\gamma_t$ is the time-dependent annealing
coefficient, and $\rho_i^\star$ controls the particle-specific
perturbation strength.
% The
% perturbation is active during the early part of the reverse process, when the global image structure is determined, and is progressively removed afterward so that the final denoising steps recover fidelity to the original condition.

\paragraph{Scheduled perturbation.}

Let $S>E$ denote the perturbation window in reverse-diffusion timestep coordinates. We define the annealing schedule as a hard-sigmoid raised to a power $p>0$, namely 
$$\gamma(t;S, E) = \max\left(0, \min\left(1, \frac{t - E}{S -E}\right)\right)^p.
$$

% \begin{equation}
% \alpha(t;S,E)
% =
% \begin{cases}
% 1,
% & t \geq S, \\[2mm]
% \dfrac{t-E}{S-E},
% & E < t < S, \\[2mm]
% 0,
% & t \leq E,
% \end{cases}
% \qquad
% \gamma(t;S,E,p)
% =
% \alpha(t;S,E)^p.
% \label{eq:perturbation_schedule}
% \end{equation}
The perturbation is fully active at the beginning of the reverse process, progressively annealed between $S$ and $E$, and completely removed afterward. For particle $i$, the condition used at timestep $t$ is
 $c_{i,t} = c_0 + \rho_i^\star\gamma(t;S,E,p)\,\Delta c_i$
where $\Delta c_i$ is its assigned pullback direction.

\paragraph{Adaptive pullback refresh.}
The pullback basis computed at the reference timestep $t_b$ only captures the
local geometry around the current state. Since this geometry evolves along the
reverse trajectory, we update the basis at $P$ intermediate timesteps:
\[
t_r^{\mathrm{ref}}
\approx
S+\frac{r}{P+1}(E-S),
\qquad
r=1,\ldots,P.
\]
At each refresh timestep, we recompute the dominant eigenspace of the local
pullback metric using the current latent of the anchor particle, and reproject each persistent Gaussian field $\eta_i$ onto its refreshed
particle-specific subspace. The previous and refreshed directions are interpolated over a short transition
window to avoid abrupt changes in the conditioning trajectory \ref{eq : pullback_refresh}.

\paragraph{Particle-specific scale selection.}
The sensitivity of the condition to response map varies along the diffusion
landscape. Consequently, a single perturbation scale may be too weak for some
particles and too strong for others. We therefore
select a particle's specific scale $\rho_i^*$ from a finite candidate set
$\mathcal R=\{\rho_1,\ldots,\rho_M\}$ by using the selection method in \ref{sec:particle_specific_scale}.

\section{Experiments \& Results}
\label{sec:experiments}

% We evaluate our approach on two conditional generation tasks: complete
% text-to-image generation and text-guided image inpainting. Our objective is
% to increase the diversity of samples generated from the same condition while
% preserving prompt alignment and visual plausibility.
For text-to-image generation, we use captions from the MSCOCO 2017
validation~\citep{lin2014microsoft}. For image inpainting, we use the BrushBench
validation set introduced with BrushNet~\citep{ju2024brushnet}.  We generate using the Stable Diffusion family of models, comparing our pullback method against baselines of DDIM, CADS~\citep{sadat2024cads}, and TPSO~\citep{meng2025training} with standard metrics of prompt alignment and diversity.  Full details are given in Section~\ref{sec:appendix-exp-details}.  
Tables~\ref{tab:coco_results} and~\ref{tab:inpainting_backbones}
summarize the results for image generation and
text-guided inpainting, respectively. Across both tasks, pullback-guided
sampling consistently improves diversity over clean DDIM while
maintaining a high level of alignment with the text condition.  
The results are particularly good for image inpainting, where the editable region constrains the possible variations and therefore makes the pullback metric a more significant initial signal. 

% \subsection{Qualitative Results}

% Figure~\ref{fig:inpainting_qualitative} presents a representative inpainting
% example. Clean DDIM generates highly consistent samples, with only minor changes details in the mask region. CADS and TPSO introduce additional variation, but
% the overall composition remain similar across particles.

% In contrast, the pullback-guided samples exhibit clearer changes in the editable area with diverse fine-grained details. This behavior is consistent with the
% regional construction of the pullback operator: conditioning directions are
% selected according to their response inside the editable region, while the
% standard inpainting blend restores the source content outside the mask.

\begin{figure}[t]
    \centering
    \includegraphics[width=\linewidth]{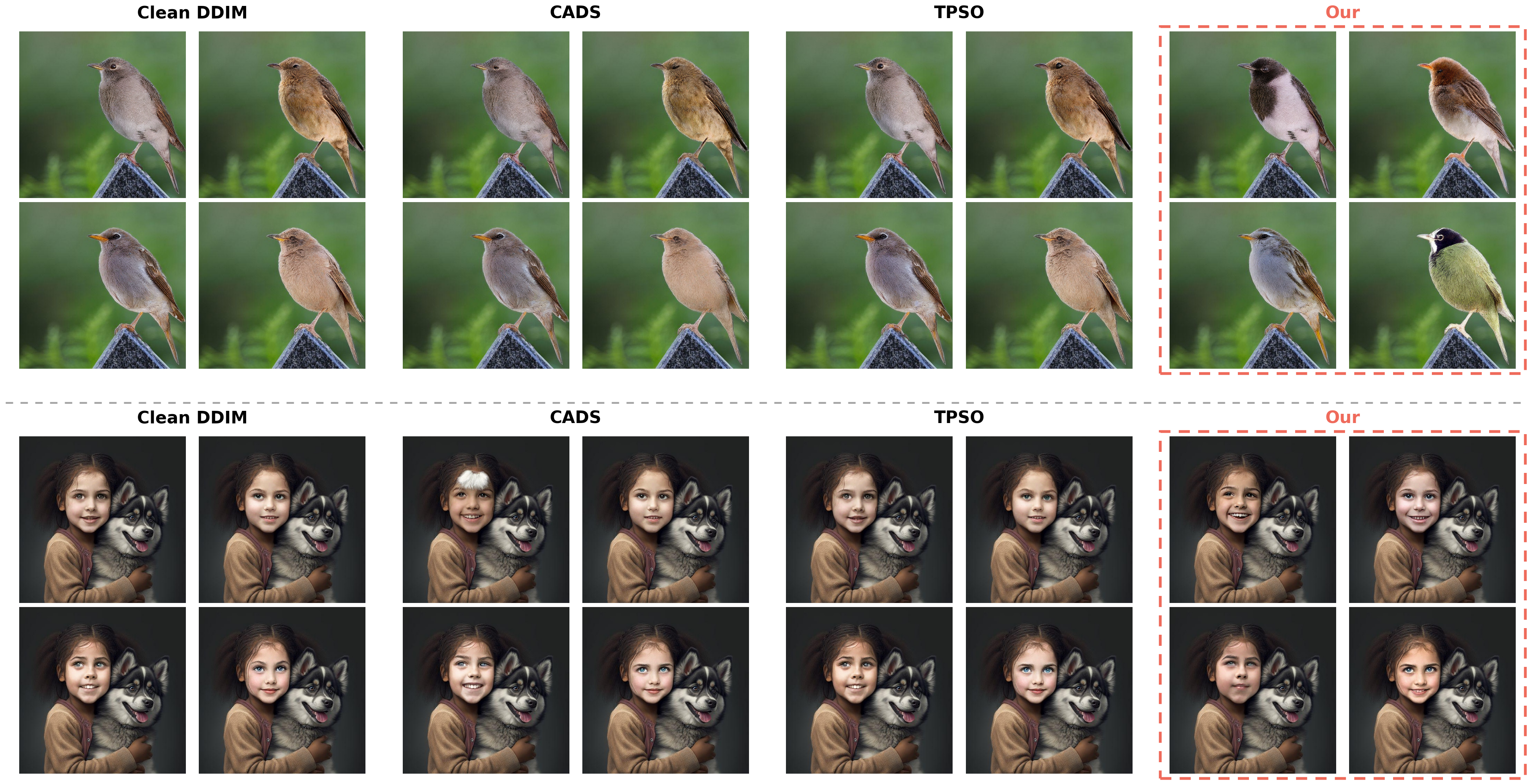}
    \caption{
    Qualitative comparison on a BrushBench example with BrushNet-SDXL. Each row contains masked samples (see Figure~\ref{fig:inpainting_qualitative_source} in the Appendix). Clean DDIM,
    CADS, and TPSO introduce local changes, whereas adaptive
    pullback produces more pronounced variations.
    }
    \label{fig:inpainting_qualitative}
\end{figure}

% For complete generation with Stable Diffusion~1.5, Pullback improves
% all three diversity metrics relative to clean DDIM. TPSO nevertheless
% achieves the strongest diversity results for this backbone. Pullback
% provides a more balanced trade-off, retaining a higher CLIP score than
% CADS and TPSO and remaining close to the clean DDIM reference.

% The diversity gains are more pronounced with Stable Diffusion XL.
% Pullback obtains the best MSS, Vendi, and DINO results among all
% evaluated methods. These improvements are accompanied by a reduction in the
% mean CLIP score from $32.0012$ for clean DDIM to $31.4248$, revealing
% the expected trade-off between stronger diversity and prompt alignment.

% The results are particularly good for image inpainting, where
% the editable region constrains the possible variations. Pullback
% achieves the best diversity values for all three metrics with both
% Stable Diffusion~1.5 and Stable Diffusion XL. 

% Overall, these results indicate that the proposed
% method is especially effective for constrained generation, where it
% produces more diverse outputs without substantially weakening
% consistency with the conditioning signal.

\begin{table*}[t]
\footnotesize
\centering
\caption{
Generation results on MSCOCO 2017 validation captions with five generated
images per caption. Stable Diffusion 1.5  and XL results use all 500 captions. Entries report mean $\pm$ standard deviation across captions.
The best mean within each backbone is shown in bold. 
}
\label{tab:coco_results}

\renewcommand{\arraystretch}{1.15}
\setlength{\tabcolsep}{8pt}

\begin{tabular}{|l|c|c|c|c|}
\hline
&
\multicolumn{1}{c|}{\textbf{Alignment}}
&
\multicolumn{3}{c|}{\textbf{Diversity}} \\
\hline

\textbf{Method}
& \textbf{CLIP $\uparrow$}
& \textbf{MSS $\downarrow$}
& \textbf{Vendi $\uparrow$}
& \textbf{DINO Mean $\downarrow$} \\
\hline

\multicolumn{5}{|l|}{
    \textit{Stable Diffusion 1.5 }
} \\
\hline

Clean DDIM
& $\mathbf{31.3957}\pm2.4730$
& $0.3359\pm0.0519$
& $4.6831\pm0.1971$
& $0.5945\pm0.1638$ \\
\hline

CADS
& $31.2537\pm2.4253$
& $0.3066\pm0.0433$
& $4.7833\pm0.1426$
& $0.5469\pm0.1628$ \\
\hline

TPSO
& $30.9592\pm2.2941$
& $\mathbf{0.3048}\pm0.0430$
& $\mathbf{4.7838}\pm0.1464$
& $\mathbf{0.5143}\pm0.1729$ \\
\hline

Pullback (ours)
& $31.3153\pm2.4671$
& $0.3194\pm0.0501$
& $4.7377\pm0.1833$
& $0.5663\pm0.1661$ \\
\hline

\multicolumn{5}{|l|}{
    \textit{Stable Diffusion XL}
} \\
\hline
Clean DDIM
& $\mathbf{32.0012}\pm2.7271$
& $0.3827\pm0.0583$
& $4.4979\pm0.2537$
& $0.6899\pm0.1571$ \\
\hline
CADS
& $31.7740\pm2.5822$
& $0.3485\pm0.0532$
& $4.6320\pm0.2089$
& $0.6374\pm0.1667$ \\
\hline
TPSO
& $31.7017\pm2.5535$
& $0.3517\pm0.0546$
& $4.6141\pm0.2192$
& $0.6263\pm0.1728$ \\
\hline
Pullback (ours)
& $31.4248\pm2.5511$
& $\mathbf{0.3196}\pm0.0470$
& $\mathbf{4.7309}\pm0.1636$
& $\mathbf{0.5639}\pm0.1785$ \\
\hline

\end{tabular}
\end{table*}
\begin{table*}[t]
\footnotesize
\centering
\caption{
BrushBench inpainting results for Stable Diffusion 1.5 and Stable
Diffusion XL, with five images per example.
Stable Diffusion 1.5 and XL results use all 500 examples.
Entries report mean $\pm$ standard deviation across examples.
The best result within each backbone is shown in bold.
}
\label{tab:inpainting_backbones}

\renewcommand{\arraystretch}{1.15}
\setlength{\tabcolsep}{8pt}

\begin{tabular}{|l|c|c|c|c|}
\hline
&
\multicolumn{1}{c|}{\textbf{Alignment}}
&
\multicolumn{3}{c|}{\textbf{Diversity}} \\
\hline

\textbf{Method}
& \textbf{CLIP $\uparrow$}
& \textbf{DINO $\downarrow$}
& \textbf{MSS $\downarrow$}
& \textbf{Vendi $\uparrow$} \\
\hline

\multicolumn{5}{|l|}{\textit{Stable Diffusion 1.5}} \\
\hline

Clean DDIM
& $\mathbf{32.0003}\pm2.7383$
& $0.8930\pm0.1094$
& $0.9643\pm0.0323$
& $1.1746\pm0.1441$ \\
\hline

CADS
& $31.9284\pm2.7278$
& $0.8736\pm0.1215$
& $0.9594\pm0.0367$
& $1.1973\pm0.1642$ \\
\hline

TPSO
& $31.9825\pm2.7077$
& $0.8807\pm0.1174$
& $0.9621\pm0.0336$
& $1.1844\pm0.1490$ \\
\hline

Pullback (ours)
& $31.8796\pm2.8055$
& $\mathbf{0.8504}\pm0.1336$
& $\mathbf{0.9498}\pm0.0414$
& $\mathbf{1.2380}\pm0.1817$ \\
\hline

% \multicolumn{5}{|l|}{
%     \textit{Stable Diffusion XL (interim, 355/500 examples)}
% } \\
% \hline

% Clean DDIM
% & $\mathbf{32.1302}\pm2.8373$
% & $0.9089\pm0.1014$
% & $0.9693\pm0.0309$
% & $1.1527\pm0.1377$ \\
% \hline

% CADS
% & $31.8247\pm2.8155$
% & $0.8813\pm0.1247$
% & $0.9618\pm0.0357$
% & $1.1854\pm0.1571$ \\
% \hline

% TPSO
% & $32.0926\pm2.8246$
% & $0.8983\pm0.1106$
% & $0.9665\pm0.0342$
% & $1.1650\pm0.1522$ \\
% \hline

% Pullback (ours)
% & $32.0007\pm2.8592$
% & $\mathbf{0.8630}\pm0.1362$
% & $\mathbf{0.9557}\pm0.0376$
% & $\mathbf{1.2124}\pm0.1657$ \\
% \hline

\multicolumn{5}{|l|}{
    \textit{Stable Diffusion XL}
} \\
\hline

Clean DDIM
& $\mathbf{32.2147}\pm2.7413$
& $0.9077\pm0.1001$
& $0.9701\pm0.0285$
& $1.1488\pm0.1273$ \\
\hline

CADS
& $32.0407\pm2.7273$
& $0.8920\pm0.1102$
& $0.9656\pm0.0308$
& $1.1692\pm0.1373$ \\
\hline

TPSO
& $32.1414\pm2.7285$
& $0.9014\pm0.1043$
& $0.9679\pm0.0306$
& $1.1585\pm0.1372$ \\
\hline

Pullback (ours)
& $32.1108\pm2.7274$
& $\mathbf{0.8788}\pm0.1246$
& $\mathbf{0.9604}\pm0.0343$
& $\mathbf{1.1918}\pm0.1519$ \\
\hline

\end{tabular}
\end{table*}

\section{Conclusion}

We introduced a training-free method for diversity
conditional diffusion models through pullback-informed prompt perturbations. Our method identifies locally expressive directions in
the conditioning space that induce the largest variation in the image space, by estimating the dominant directions
of the pullback operator.  These results demonstrate that the local geometry of the denoiser with respect to condition provides useful directions for diversifying generation without fine-tuning. Future work will evaluate the effectiveness of pullback directions on other constrained generation tasks.

% Experiments on complete text-to-image generation and image inpainting
% show that pullback-guided sampling consistently improves diversity over
% clean DDIM while preserving strong alignment with the input condition.
% The method achieves the strongest diversity results across all reported
% metrics for Stable Diffusion XL generation and for both inpainting
% backbones. For Stable Diffusion~1.5, TPSO obtains the highest
% diversity for full image generation, while Pullback provides a competitive compromise between
% diversity and prompt alignment for both full image generation and image inpainting. The qualitative examples further show
% that the method produces meaningful variations inside editable regions
% while preserving the surrounding image context.

\newpage
\bibliography{sample}
\newpage
%\clearpage
\appendix

\section{Related works}

\subsection{Image Diffusion Models}
\paragraph{Conditional image generation}
Conditional image generation aims to synthesize realistic images that satisfy user-specified controls. These conditions may take the form of textual descriptions, reference images, class labels, segmentation maps, depth maps, or spatial layouts. Diffusion models have become a dominant framework for this task because of their strong image quality and flexibility in incorporating heterogeneous conditions. Text-conditioned latent diffusion models introduce semantic information through cross-attention, enabling high-resolution text-to-image synthesis \citep{rombach2022high}, while classifier-free guidance strengthens alignment by combining conditional and unconditional denoising predictions \citep{ho2022classifier}. Other approaches introduce image-to-image conditioning \citep{saharia2021palette} or additional spatial controls such as edges, poses, depth maps, and layouts \citep{zheng2023layoutdiffusion,zhang2023adding}. These developments have substantially improved the controllability and fidelity of conditional image synthesis.
\paragraph{Image inpainting}
Image inpainting consists in reconstructing a missing region of an image while preserving the visible content and maintaining semantic and visual coherence across the edited region. Diffusion-based approaches address inpainting either by training models that explicitly receive the masked image and mask as conditioning inputs or by adapting the reverse process of a pretrained generative model \citep{ju2024brushnet,lugmayr2022repaint, nichol2021glide}. These approaches aim to generate plausible content within the masked region while preserving the observed pixels and maintaining semantic and visual coherence with the unmasked region.

\paragraph{Diversity image generation}
Although diffusion models generate samples from random initial noise, stochasticity alone does not guarantee meaningful diversity under a fixed condition. Different initializations may still converge toward semantically similar outputs, particularly when strong guidance, textual constraints, or spatial observations restrict the reverse process. Recent analyses therefore emphasize a trade-off between realism, condition consistency, and distributional coverage \citep{astolfi2024consistency}, while other work shows that the initial noise can already strongly determine the spatial organization of the final image \citep{ban2025crystal}. Consequently, producing multiple distinct yet equally valid completions remains challenging, especially for constrained tasks such as text-guided inpainting. Existing methods address this limitation through additional training(finetuning diversity methods), condition perturbations, or interactions between sampling trajectories.
\subsection{Fine-tuning diversity methods}

Fine-tuning methods enable further training towards a target reward or distributional loss.  For a scalar reward, Adjoint Matching~\citep{domingo2025adjoint} backpropagates through the entire dynamics to improve the reward while staying close to the original reverse process.  This has been generalized as a tool to promote diversity in the generated distribution in ~\citet{de2025provable}.  Notably, these methods require expensive sampling through the diffusion model and are therefore no longer simulation-free.

\subsection{Inference-Time Diversity Methods}

Inference-time diversity methods aim to improve the coverage of the conditional
distribution without retraining the underlying diffusion model. Existing
approaches differ primarily in the component of the sampling process on which
they intervene. Condition-space methods modify the conditioning signal supplied
to the denoising network. CADS \citep{sadat2024cads} perturbs the condition embedding with Gaussian
noise whose magnitude is progressively annealed throughout sampling, thereby
increasing variation during the early stages of generation. TPSO \citep{meng2025training} instead explores underutilized directions
of the token-embedding space through optimized prompt perturbations, while
constraining the resulting samples to preserve semantic consistency. Some family directly modifies the source of stochasticity by optimizing the
initial diffusion noise \citep{harrington2025s,eyring2024reno} , an others instead construct a joint,
non-i.i.d. distribution sample \citep{corso2024particle}. Particle Guidance introduces pairwise repulsive interactions during denoising
to prevent particles from converging toward the same modes \citep{corso2024particle}.

\section{Pullback-Informed Prompt Perturbation}
\label{app:pullback_algorithm}

Let $C_0^+\in\mathbb{R}^{L\times d_c}$ denote the
positive token condition, where
$d_c$ is the token-embedding dimension. After vectorization, the
condition belongs to $\mathbb{R}^D$, with
$D=Ld_c$.
\subsection{Matrix-free estimation of expressive directions}\label{sec:approximate-svd}

We first follow the clean CFG-DDIM trajectory of one reference particle
to a basis-estimation timestep $t_b$, obtaining $z_{t_b}^{\mathrm{ref}}$.
At the fixed state $(z_{t_b}^{\mathrm{ref}},t_b,C_0^+)$, we seek
directions in the conditioning space that induce the largest local
variation in the denoiser output. For a unit direction $v$, this
variation can be measured by
\[
    \frac{1}{\varepsilon^2}
    \left\|
        f_b(C_0^+ + \varepsilon v)-f_b(C_0^+)
    \right\|_2^2,
    \qquad \|v\|_2=1,
\]
where
$f_b(C)=\epsilon_\theta(z_{t_b}^{\mathrm{ref}},t_b,C). $
Using the first-order approximation
$f_b(C_0^++\varepsilon v)-f_b(C_0^+)\simeq\varepsilon J_bv$,
the problem of finding the $K$ most expressive orthonormal directions
becomes

\begin{equation}
\label{eq:conditional_pullback_norm}
    V^\star
    =
    \arg\max_{V^\top V=I_K}
    \|J_bV\|_F^2
    =
    \arg\max_{V^\top V=I_K}
    \operatorname{tr}\!\left(V^\top J_b^\top J_bV\right).
\end{equation}

Therefore, by the Rayleigh--Ritz theorem, the columns of $V^\star$
are the $K$ dominant eigenvectors of
\[
    G_b=J_b^\top J_b,
    \qquad
    J_b=\left.\frac{\partial f_b}{\partial C}\right|_{C=C_0^+}.
\]
Equivalently, they are the dominant right singular vectors of $J_b$.
We estimate these directions without explicitly constructing $J_b$.
% At the fixed state $(z_{t_b}^{\mathrm{ref}},t_b,C_0^+)$, define
% \[
% f_b(C)=\epsilon_\theta(z_{t_b}^{\mathrm{ref}},t_b,C),
% \qquad
% G_b=J_b^\top J_b,
% \qquad
% J_b=\frac{\partial f_b}{\partial C}.
% \]

% The operator is applied without constructing $J_b$. For a direction $v$,
\[
J_bv
\approx
\frac{f_b(C_0^++\varepsilon v)-f_b(C_0^+-\varepsilon v)}
     {2\varepsilon},
\qquad
G_bv=J_b^\top(J_bv),
\]
where the second operation is evaluated with a vector--Jacobian product.

To approximate the $K$ dominant eigenvectors, we use randomized block
subspace iteration followed by Rayleigh--Ritz extraction
\citep{halko2011finding}. We sample
$\Omega\in\mathbb R^{D\times K}$ with Gaussian entries and set
\[
Q_0=\operatorname{QR}(\Omega),\qquad
Q_{r+1}=\operatorname{QR}(G_bQ_r),\quad r=0,\ldots,R-1.
\]
We then construct
\[
B=\frac{1}{2}\left(Q_R^\top G_bQ_R+
                         (Q_R^\top G_bQ_R)^\top\right),
\]
diagonalize $B=U\Lambda U^\top$, sort the eigenvalues in decreasing
order, and lift the resulting vectors back to condition space:
\[
V=Q_RU=[v_1,\ldots,v_K].
\]

\subsection{Fixed disjoint perturbation directions}

The basis indices are divided into balanced, mutually disjoint sets
$\mathcal I_1,\ldots,\mathcal I_N$. For each particle, a Gaussian field
$\xi_i\sim\mathcal N(0,I_D)$ is sampled once and kept fixed. Its
projection onto the assigned pullback subspace is
\[
\bar d_i
=
\sum_{k\in\mathcal I_i}
\langle \xi_i,v_k\rangle v_k,
\qquad
d_i=\frac{\bar d_i}{\|\bar d_i\|_2}.
\]

To express the perturbation relative to the magnitude of the original
condition, each real token is scaled by its clean embedding norm:
\begin{equation}
\Delta c_i
=||C_0^+||_2d_i,
\label{eq:delta_ci}
\end{equation}
where $||C_0^+||_2$ denote the norms of the token embeddings. 
\paragraph{Adaptive pullback refresh}
For the refresh, the pullback basis and the corresponding directions
are recomputed at intermediate denoising steps. Let $d_i^{\mathrm{old}}$ and
$d_i^{\mathrm{new}}$ denote the directions before and after recomputation.
To avoid an abrupt change, we interpolate between them over $L=2$ steps:
\begin{equation}
d_i^{(\ell)}
=
\frac{
(1-\beta_\ell)d_i^{\mathrm{old}}
+\beta_\ell d_i^{\mathrm{new}}
}{
\left\|
(1-\beta_\ell)d_i^{\mathrm{old}}
+\beta_\ell d_i^{\mathrm{new}}
\right\|_2
},
\qquad
\beta_\ell=\frac{\ell}{2},
\quad \ell\in\{1,2\}.
\label{eq : pullback_refresh}
\end{equation}.

\subsection{Particle-specific magnitude selection}
\label{sec:particle_specific_scale}
We select a particle's specific
magnitude from a finite candidate set
$\mathcal R=\{\rho_1,\ldots,\rho_M\}$ at a specific timestep $t_p$ before the final sampling run. 

For each particle $i$ and candidate magnitude $\rho_m$, we compute the
predicted clean image $\widehat{x}_{i,m}$ at $t_p$. We additionally use
$\rho_0=0$ to obtain the unperturbed reference $\widehat{x}_{i,0}$.
Semantic diversity between two candidates is measured using the cosine
distance between their DINO representations:
\begin{equation}
d_{\mathrm{DINO}}(x,x')
=
1-
\cos\left(
\phi_{\mathrm{DINO}}(x),
\phi_{\mathrm{DINO}}(x')
\right),
\label{eq:dino_distance}
\end{equation}
where $\phi_{\mathrm{DINO}}$ is a frozen DINO image encoder. Prompt alignment
is measured by
\begin{equation}
s_{\mathrm{CLIP}}(x,C)
=
\cos\left(
\phi_{\mathrm{CLIP}}^{I}(x),
\phi_{\mathrm{CLIP}}^{T}(C)
\right),
\label{eq:clip_alignment}
\end{equation}
where $\phi_{\mathrm{CLIP}}^{I}$ and $\phi_{\mathrm{CLIP}}^{T}$ are the
frozen CLIP image and text encoders, respectively.

The particle-specific magnitudes are jointly selected by maximizing the
average pairwise DINO distance while limiting the loss of prompt alignment
relative to the unperturbed prediction:
\begin{equation}
\begin{aligned}
(m_1^\star,\ldots,m_B^\star)
\in
\arg\max_{\substack{m_1,\ldots,m_B\\m_i\in\{1,\ldots,M\}}}
\quad&
\frac{2}{B(B-1)}
\sum_{i<j}
d_{\mathrm{DINO}}
\left(
\widehat{x}_{i,m_i},
\widehat{x}_{j,m_j}
\right)
\\
\text{subject to}\quad&
s_{\mathrm{CLIP}}
\left(
\widehat{x}_{i,m_i},C
\right)
\geq
s_{\mathrm{CLIP}}
\left(
\widehat{x}_{i,0},C
\right)
-
\delta_{\mathrm{CLIP}},
\qquad i=1,\ldots,B.
\end{aligned}
\label{eq:particle_specific_scale_selection}
\end{equation}

The selected magnitudes are
$\rho_i^\star=\rho_{m_i^\star}$.

\subsection{Scheduled reverse sampling}

After selecting $\rho_i^\star$, sampling is restarted from the original
initial latents. The perturbation envelope is
\[
\alpha(t)=
\begin{cases}
1, & t\geq T_S,\\[2mm]
\left(\dfrac{t-T_E}{T_S-T_E}\right)^p,
   & T_E<t<T_S,\\[3mm]
0, & t\leq T_E.
\end{cases}
\]
At each reverse step, the particle-specific positive condition is
\[
C_{i,t}^+
=
C_0^+
+
\alpha(t)\rho_i^\star \Delta_{Ci,t}
\qquad
C_{i,t}^-=C_0^-.
\]
The resulting pair is used in the standard CFG prediction, followed by
one DDIM update.

\section{Experimental Details}\label{sec:appendix-exp-details}

\paragraph{Tasks and datasets}
For complete text-to-image generation, we use captions from the MSCOCO 2017
validation~\citep{lin2014microsoft}. For image inpainting, we use the BrushBench
validation set introduced with BrushNet~\citep{ju2024brushnet}. Each inpainting
example contains a source image, a binary edit mask, and a text condition
describing the expected content.
We evaluate 500 conditions from each benchmark and generate five images per
condition, resulting in 2,500 generated images for every method on each task.

\paragraph{Configuration}
Complete generation is performed with Stable Diffusion~1.5. The inpainting
evaluation uses the corresponding Stable Diffusion~1.5 and XL BrushNet
backbone~\citep{ju2024brushnet}. We use 50 DDIM
sampling steps, classifier-free guidance with scale $7.5$, and $\eta=0$.
We follow the publicly available default configurations of the corresponding
models.

For every condition, all compared methods receive exactly the same initial
latent particles. All experiments use fixed random seeds to ensure fair and model-consistent comparisons.

\paragraph{Compared methods}
We compare the following sampling procedures:
\begin{itemize}
    \item \textbf{Clean DDIM}: standard DDIM sampling without condition
    perturbation;
    \item \textbf{CADS}: condition annealing with isotropic Gaussian
    perturbations~\citep{sadat2024cads};
    \item \textbf{TPSO}: token-prompt embedding space optimization
    applied during sampling~\citep{meng2025training};
    \item \textbf{Pullback}: our method, which perturbs the condition along
    dominant pullback directions and uses particle-specific perturbation
    magnitudes.
\end{itemize}

\subsection{Evaluation Metrics}

The evaluation measures three complementary properties: semantic alignment,
perceptual diversity, and set-level diversity.

\paragraph{Text--image alignment.}
Text–image consistency is evaluated
using the CLIP Score~\citep{hessel2021clipscore}, defined as the cosine similarity
between CLIP embeddings of the generated image and its
corresponding prompt. Higher values indicate stronger semantic alignment with the prompt.

\paragraph{Semantic diversity.}
We extract DINOv2 image representations ~\citep{oquab2023dinov2} and compute
the mean pairwise cosine similarity within each set of samples:
Lower mean similarity indicates greater overall semantic diversity. 

\paragraph{Set-level diversity.}
We report the Mean Similarity Score (MSS) and Vendi
Score~\citep{friedman2022vendi} for per-prompt (intra-set) diversity, both computed using SSCD features following \citep{sadat2024cads}. Lower MSS and higher Vendi indicates greater within-condition diversity.

\begin{figure}[t]
    \centering

    % Left side: image
    \begin{minipage}[t]{0.62\linewidth}
        \vspace{0pt}
        \centering
        \includegraphics[width=\linewidth]
        {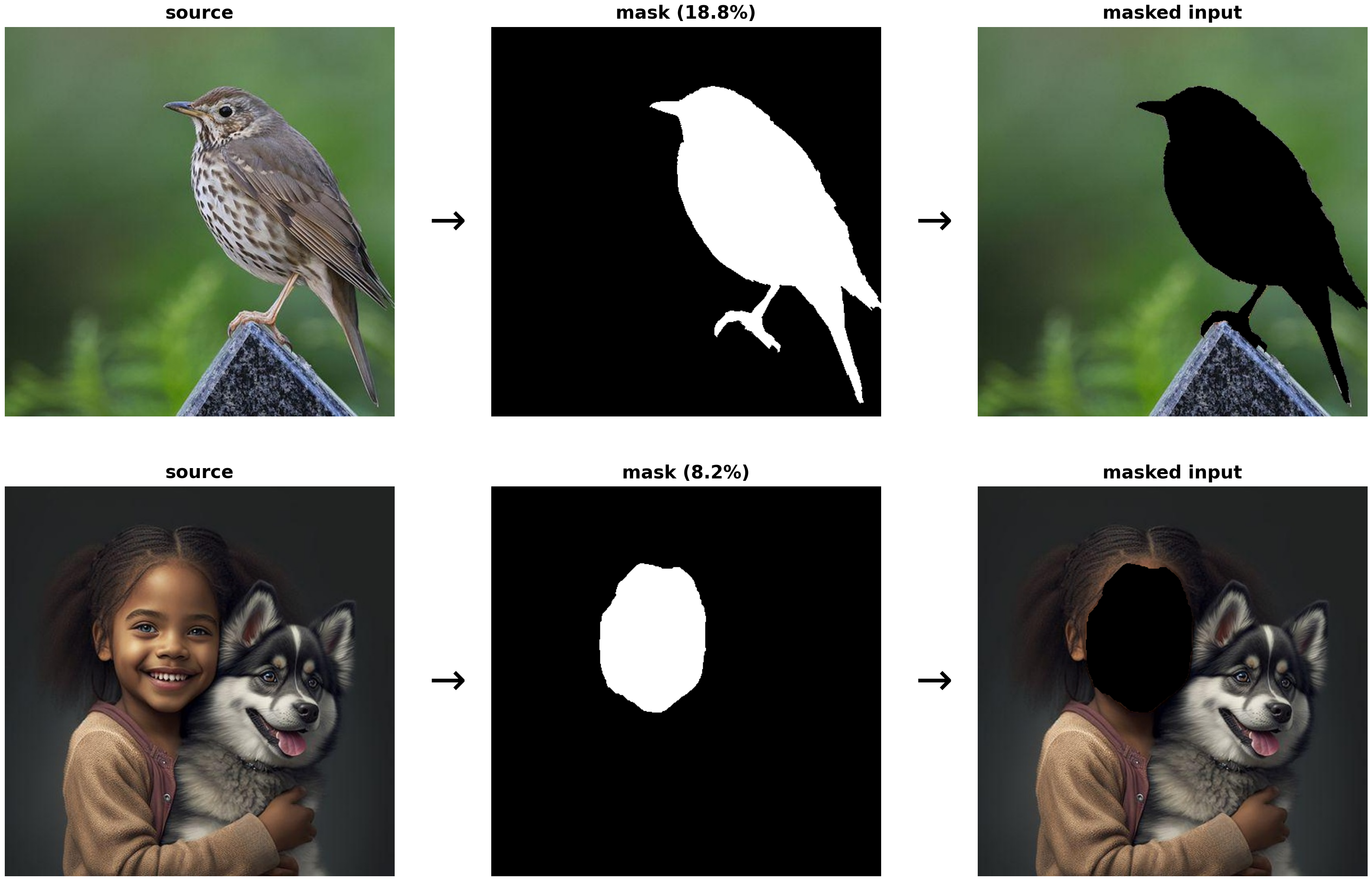}
    \end{minipage}
    \hfill
    % Right side: description
    \begin{minipage}[t]{0.34\linewidth}
        \vspace{0pt}
        \small
        \caption{
            Illustration of the text-guided inpainting task.
            Given a text prompt, the model regenerates the masked
            region while preserving the unmasked region.
        }
        The prompts are \emph{``a small bird perched on top of a
        rock''} (top) and \emph{``a young girl holding a husky dog
        on a black background''} (bottom).
        \label{fig:inpainting_qualitative_source}
    \end{minipage}

\end{figure}

\begin{figure}[t]
    \centering

    \begin{minipage}[c]{0.64\linewidth}
        \centering
        % Crop the first two panels and the global title.
        \includegraphics[
            width=\linewidth,
        ]{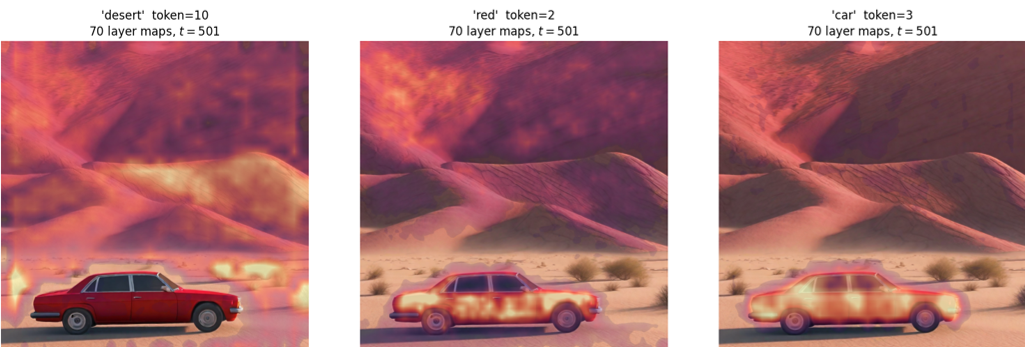}
    \end{minipage}
    \hfill
    \begin{minipage}[c]{0.32\linewidth}
        \small
        \textbf{Prompt.}\\[2mm]
        \textit{``a red car with tinted windows driving through the desert, photorealistic, 8k''}
    \end{minipage}

    \caption{
    Token-dependent cross-attention at DDIM timestep $t=499$. Each visualization averages 28 U-Net
    cross-attention maps over attention heads and layers. Warmer regions receive greater relative attention for the
    corresponding token. 
    % The two tokens produce distinct spatial patterns:
    % ``skirt'' is more strongly associated with the article of clothing,
    % whereas ``wall'' is associated with the background. This
    % illustrates how individual components of the text condition contribute
    % differently to the spatial guidance of denoising.
    }
    \label{fig:token_cross_attention}
\end{figure}

\section{Further Experimental Details}
\subsection{Feature-space and conditioning-space pullback directions}
\label{sec:feature_condition_pullback}

Park et al.~\cite{park2023understanding} use the pullback metric to
identify locally sensitive directions in the noisy latent space of a
diffusion model. Their construction relies on the observation that the
intermediate representations of the denoising model encode semantic and
spatial information about the generated image. Let
\begin{equation}
    f_t(\cdot;c):\mathcal X_t\rightarrow\mathcal H_t
\end{equation}
denote the mapping from the noisy latent $x_t$ to a selected intermediate
U-Net feature representation $h_t=f_t(x_t;c)$ under condition $c$. Its
Jacobian and induced pullback metric are
\begin{equation}
    J_x
    =
    \frac{\partial f_t(x_t;c)}{\partial x_t},
    \qquad
    G_x
    =
    J_x^\top J_x.
\end{equation}
The dominant eigenvectors $\{v_k\}$ of $G_x$, identify input directions that produce
large local variations in the feature representation.

For direction $v_k$ and strength $s$, we construct the norm-matched
latent perturbation
\begin{equation}
    \delta x_{t,k}(s)
    =
    s
    \frac{\lVert x_t\rVert_2}{\lVert v_k\rVert_2}
    v_k.
    \label{eq:feature_probe}
\end{equation}
Following the $x$-space guidance formulation of
Park et al.~\cite{park2023understanding}, the perturbed latent is
\begin{equation}
    \widetilde{x}_{t,k}(s)
    =
    x_t
    +
    \gamma
    \left[
        \epsilon_\theta
        \bigl(x_t+\delta x_{t,k}(s),t,c\bigr)
        -
        \epsilon_\theta(x_t,t,c)
    \right],
    \label{eq:xspace_guidance}
\end{equation}
where $\gamma$ controls the magnitude of the noise-prediction response.

We compare this construction with the conditioning-space pullback. For
a fixed latent $x_t$, we consider the condition-response mapping
\begin{equation}
    g_t(C)
    =
    \epsilon_\theta(x_t,t,C),
\end{equation}
with Jacobian and pullback metric
\begin{equation}
    J_C
    =
    \frac{\partial \epsilon_\theta(x_t,t,C)}{\partial C},
    \qquad
    G_C
    =
    J_C^\top J_C.
\end{equation}
The dominant eigenvectors $\{d_k\}$ of $G_C$ identify condition
directions that induce large changes in the predicted noise.

For both constructions, we compute a rank-$4$ approximation of the
pullback metric at the same latent and retain the four dominant
directions. We keep others parameters same for both. For the latent-feature pullback, we
evaluate
\begin{equation}
    s\in
    \{0.005,0.015,0.02,0.025,0.03,0.04,0.05\},
\end{equation}
with $\gamma=5$. For the conditioning pullback, we evaluate
\begin{equation}
    s\in
    \{0.1,0.25,0.5,1.0,1.5,2.0,2.5\}.
\end{equation}
The numerical values of $s$ are not directly comparable across the two
methods because the perturbations operate in different spaces with different
sensitivity scales. We therefore select a specific range of values
that clearly reveals the progressive effect of the perturbation on the
generated image. The experiment compares how the output
evolves within each method as the perturbation strength increases.
\paragraph{Qualitative observations.}
The latent-feature pullback primarily produces localized variations and
does not induce substantial semantic diversity. Increasing the perturbation
strength does not necessarily amplify these variations in a meaningful way;
instead, it progressively introduces high-contrast contours, geometric
distortions, and other off-manifold artifacts. In contrast, the
conditioning-space pullback produces broader semantic transformations that
affect a larger portion of the image. These transformations remain
direction-dependent across the strength sweep, with different pullback
directions controlling distinct semantic attributes.

\begin{figure*}[t]
    \centering
    \includegraphics[width=\textwidth]
    {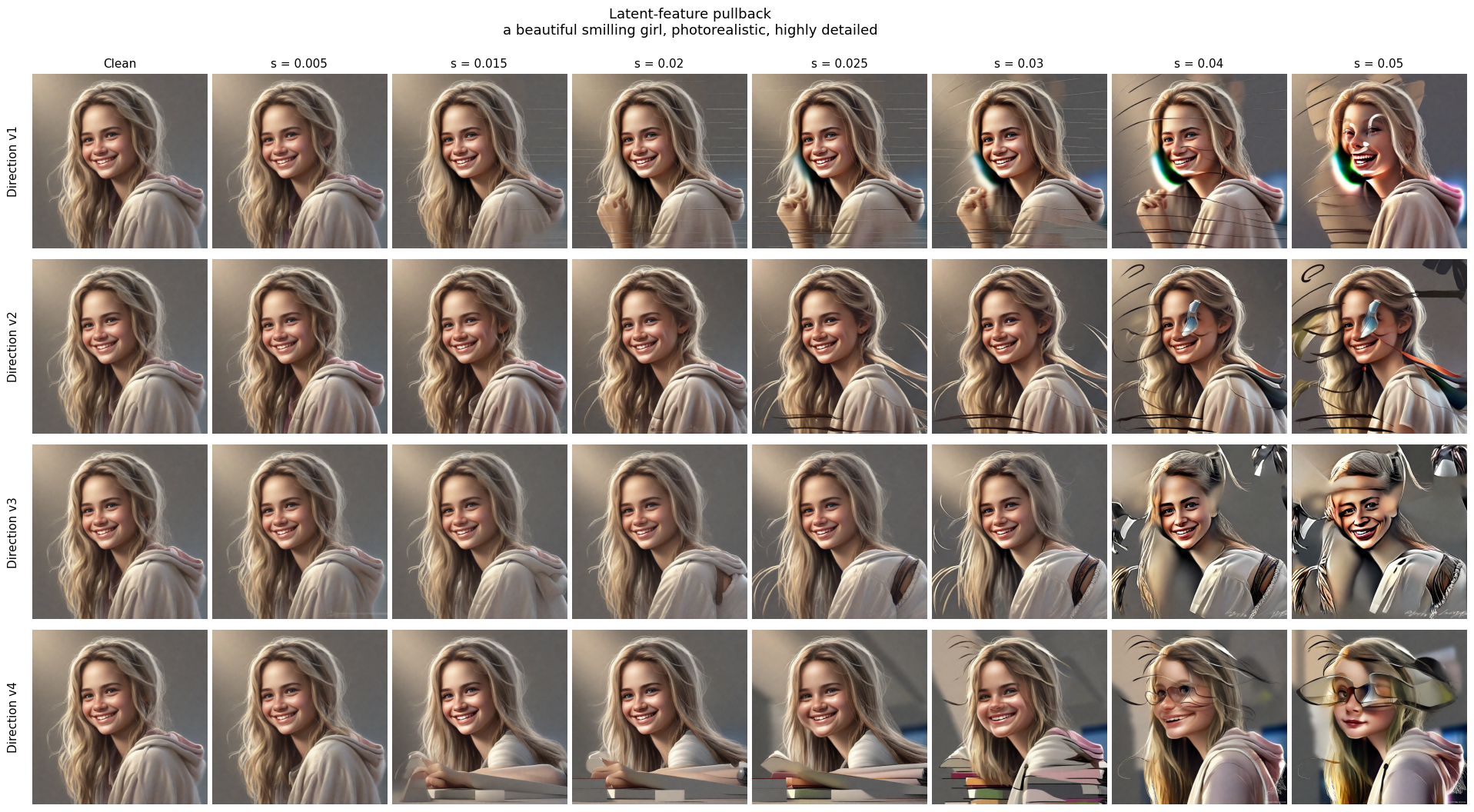}
    \caption{
        Effect of the perturbation strength along latent-feature pullback
        directions. Rows correspond to the four dominant directions
        $v_1,\ldots,v_4$ of the pullback metric
        $G_x=J_x^\top J_x$, where $J_x$ is the Jacobian of the selected
        U-Net feature representation with respect to the noisy latent
        $x_t$. Columns show the clean sample followed by increasing
        strengths.
        Moderate strengths produce predominantly localized, whereas larger strengths introduce visible distortions.
    }
    \label{fig:feature_pullback_strength}
\end{figure*}
\begin{figure*}[t]
    \centering
    \includegraphics[width=\textwidth]
    {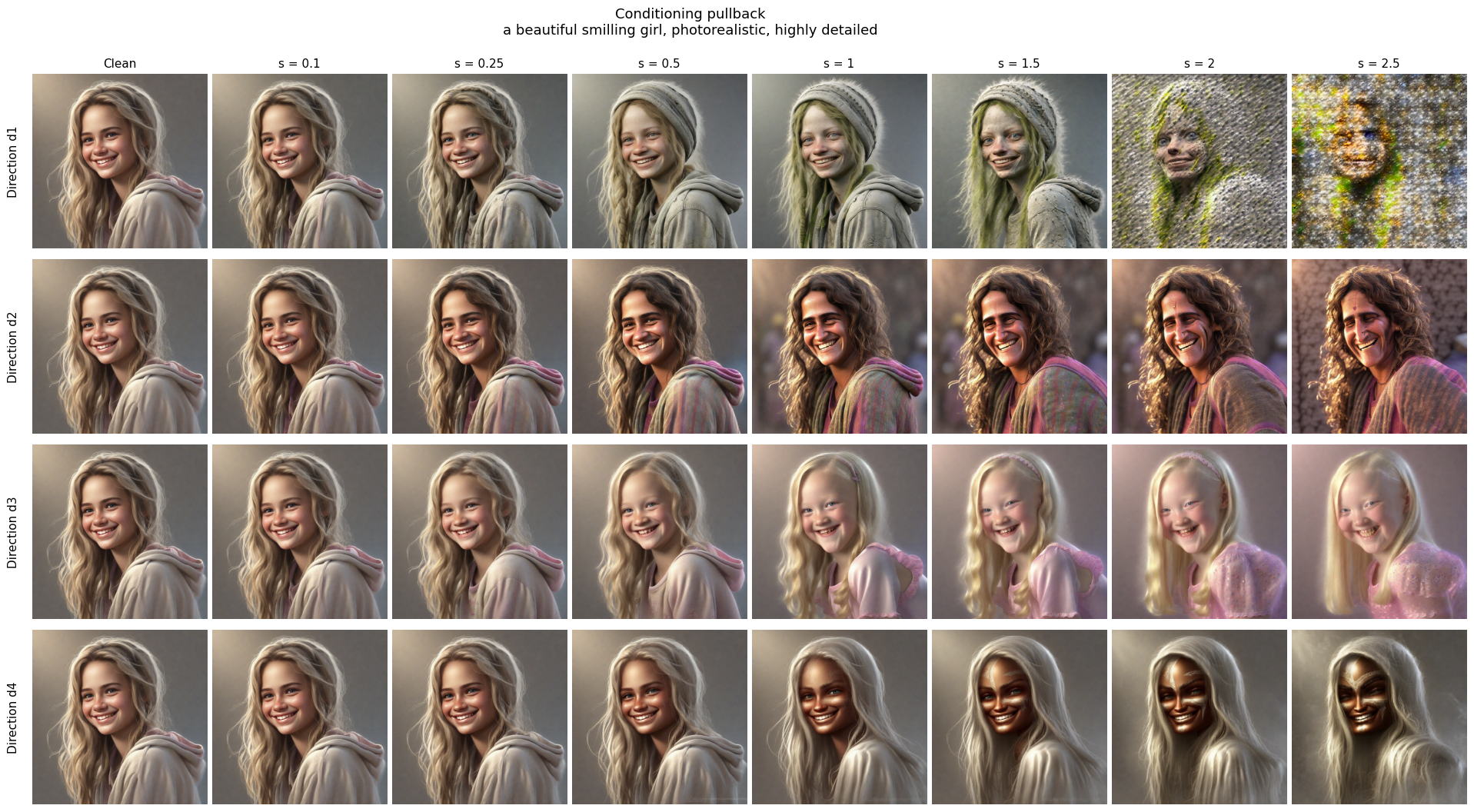}
    \caption{
        Effect of the perturbation strength along conditioning-space
        pullback directions. Rows correspond to the four dominant
        directions of the condition pullback metric
        $G_C=J_C^\top J_C$, where
        $J_C=\partial\epsilon_\theta(x_t,t,C)/\partial C$.
        Columns show the clean sample followed by increasing strengths.
        The different directions produce distinct global semantic transformations, while large strengths can reduce image quality and fidelity.
    }
    \label{fig:conditioning_pullback_strength}
\end{figure*}
\begin{figure*}[t]
    \centering
    \includegraphics[width=0.82\textwidth]
    {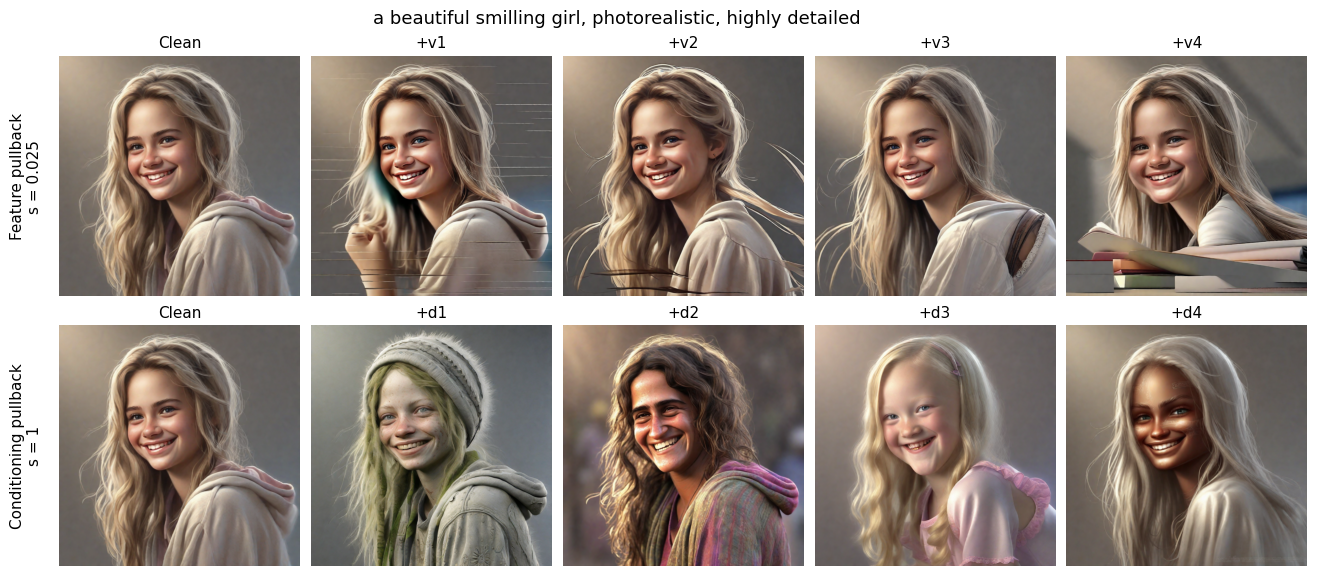}
    \caption{
        Qualitative comparison between latent-feature and
        conditioning-space pullback directions computed from the same
        anchor. The top row shows the clean sample and the four dominant
        latent-feature directions at $s=0.025$, using an
        $x$-space guidance scale of $\gamma=5$. The bottom row shows the
        clean sample and the four dominant conditioning directions at $s=1$. In this example, the latent-feature
        directions primarily affect localized appearance, pose, and nearby
        image structure, whereas the conditioning-space directions induce
        broader semantic changes.
        All results use the same prompt, initial noise, and
        reverse-process parameters.
    }
    \label{fig:feature_conditioning_comparison}
\end{figure*}
\end{document}